\documentclass[10pt,twocolumn,letterpaper]{article}
\makeatother

\usepackage[pagenumbers]{cvpr} 
\usepackage{multirow}

\usepackage[dvipsnames]{xcolor}

\definecolor{cvprblue}{rgb}{0.21,0.49,0.74}
\usepackage[pagebackref,breaklinks,colorlinks,citecolor=cvprblue]{hyperref}

\definecolor{mygray}{gray}{.9}

\def\paperID{4136} 
\def\confName{CVPR}
\def\confYear{2024}

\title{Efficient Multimodal Semantic Segmentation via Dual-Prompt Learning}

\author{Shaohua Dong$^{1}$\;\;\; Yunhe Feng$^{1}$\;\;\;  Qing Yang$^{1}$\;\;\;
Yan Huang$^{1}$\;\;\; Dongfang Liu$^{2}$\;\;\;  Heng Fan$^{1}$\\
$^1$University of North Texas \;\;\;\;\;\;\; $^2$Rochester Institute of Technology\\
}

\begin{document}
\maketitle

\begin{abstract}

Multimodal (\eg, RGB-Depth/RGB-Thermal) fusion has shown great potential for improving semantic segmentation in complex scenes (\eg, indoor/low-light conditions).
Existing approaches often fully fine-tune a dual-branch encoder-decoder framework with a complicated feature fusion strategy for achieving multimodal semantic segmentation, which is training-costly due to the massive parameter updates in feature extraction and fusion. To address this issue, we propose a surprisingly simple yet effective dual-prompt learning network (dubbed DPLNet) for \emph{training-efficient} multimodal (\eg, RGB-D/T) semantic segmentation. The core of DPLNet is to directly adapt a frozen pre-trained RGB model to multimodal semantic segmentation, reducing parameter updates. For this purpose, we present two prompt learning modules, comprising multimodal prompt generator (MPG) and multimodal feature adapter (MFA). MPG works to fuse the features from different modalities in a compact manner and is inserted from shadow to deep stages to generate the multi-level multimodal prompts that are injected into the frozen backbone, while MPG adapts prompted multimodal features in the frozen backbone for better multimodal semantic segmentation. Since both the MPG and MFA are lightweight, only a few trainable parameters (3.88M, 4.4\% of the pre-trained backbone parameters) are introduced for multimodal feature fusion and learning. Using a simple decoder (3.27M parameters), DPLNet achieves new state-of-the-art performance or is on a par with other complex approaches on four RGB-D/T semantic segmentation datasets while satisfying parameter efficiency. Moreover, we show that DPLNet is general and applicable to other multimodal tasks such as salient object detection and video semantic segmentation. Without special design, DPLNet outperforms many complicated models. Our code will be available at \href{https://github.com/ShaohuaDong2021/DPLNet}{github.com/ShaohuaDong2021/DPLNet}.

\end{abstract}

\section{Introduction}
\label{sec:intro}

Semantic segmentation, that aims at assigning each pixel in an image with one of the pre-define labels, is a fundamental problem in computer vision with a wide spectrum of crucial applications such as intelligent driving and robotics, and has seen considerable progress (\eg~\cite{segformer,chen2017deeplab,long2015fully,fan2018multi,pvt,maskformer,strudel2021segmenter}) in recent years. Despite this, the RGB-based segmentation approaches might largely degenerate when applied to complex scenarios 
(\eg, in the \emph{cluttered indoor} environment or the \emph{low-light} condition). To address this challenge, an auxiliary modality (\eg, Depth or Thermal), which provides supplemental information to the RGB image, has been used for achieving multimodal, RGB+Depth (RGB-D)~\cite{wang2016learning,SA-gate,shapeconv,CMX,CMNext} or RGB+Thermal (RGB-T)~\cite{mfnet,RTFNet,EGFNet,CMX,CMNext}, semantic segmentation, demonstrating promising performance.

\begin{figure}[t]
 \centering
 \includegraphics[width=0.94\linewidth]{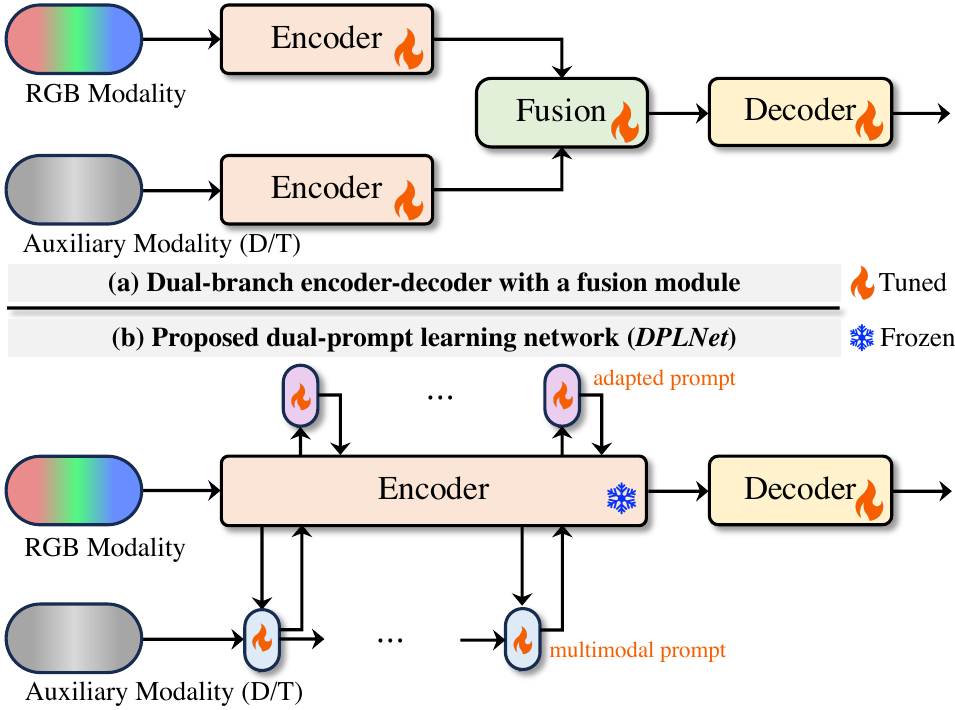}
 \caption{Compared with current dual-branch encoder-decoder architecture with a fusion module (see image (a)), the proposed dual-prompt learning paradigm (see image (b)) does not require tuning the weighty encoder pre-trained on RGB data and only introduces a few trainable parameters for multimodal feature fusion and adaption, achieving parameter-efficient training for multimodal semantic segmentation. \emph{Best viewed in pdf for all figures in this paper}.}
 \label{fig:comp}
 \vspace{-1em}
\end{figure}

\begin{figure*}[!t]
 \centering
 \includegraphics[width=0.97\linewidth]{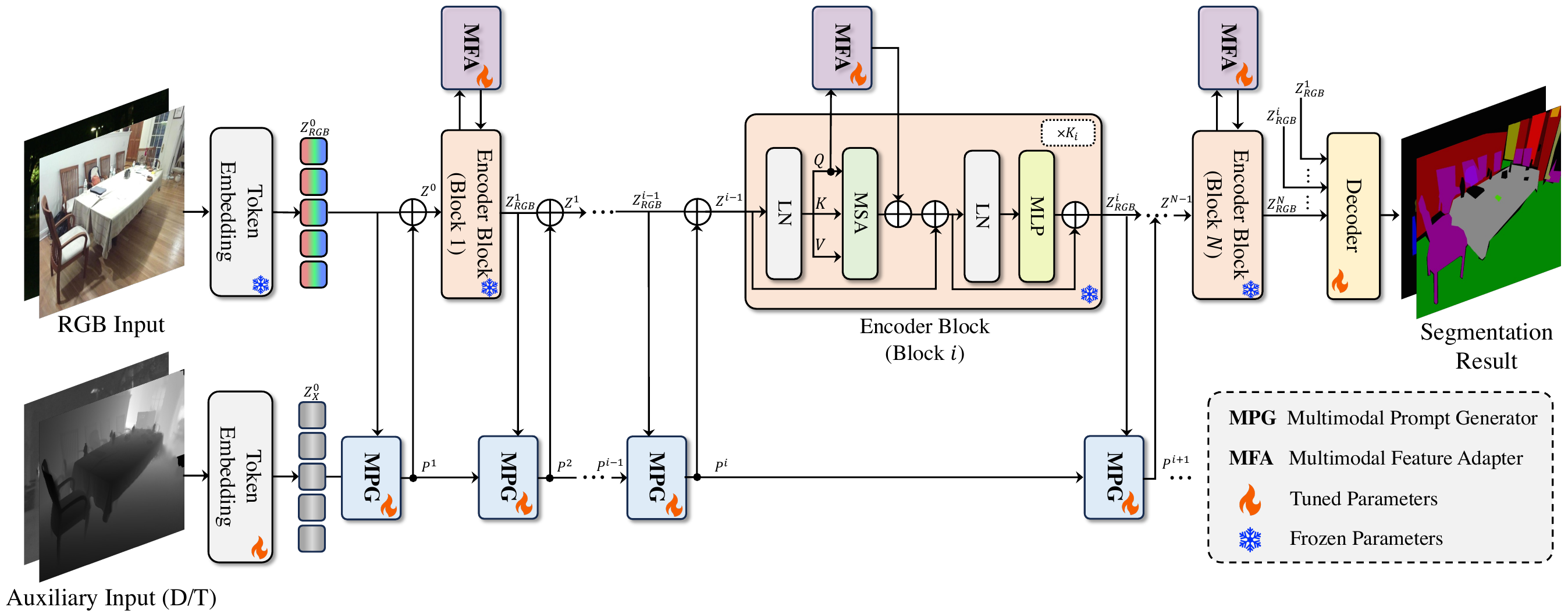}
 \caption{Overview architecture of the proposed \emph{DPLNet}, which adapts a frozen pre-trained model using two specially designed prompting learning modules, \emph{MPG} for multimodal prompt generation and \emph{MFA} for multimodal feature adaption, with only a few learnable parameters to achieve multimodal semantic segmentation in a training-efficient way.}
 \label{fig:framework}
 \vspace{-2mm}
\end{figure*}

Existing multimodal methods for semantic segmentation often adopt a straight-forward dual-branch encoder-decoder architecture (\eg,~\cite{EGFNetTits,EGFNet,GMNet, MTANet,fuseseg,CMX}), where one branch is adopted for feature extraction of the RGB modality and the other one for the feature of the auxiliary modality (see Fig.~\ref{fig:comp} (a)). Afterwards, a complex fusion strategy is employed to merge the multimodal features to achieve semantic segmentation. Albeit simple, such a framework often needs to \emph{fully fine-tune} the entire network, which is \emph{training-costly} due to massive parameter updates in the feature extraction and fusion. In addition, it requires to retain two heavy encoders after training, and thus increases the deployment burden of multimodal semantic segmentation. Moreover, the auxiliary encoder branch in existing methods is often initialized with a pre-trained RGB model (\eg,~\cite{ResNet, convnext, segformer}) and then fully fine-tuned, which may lead to suboptimal performance due to domain gap between RGB and auxiliary modalities~\cite{abmdrnet}. Noticing these issues, a question naturally arises: \emph{Is there a better way that is \textbf{training-efficient}, \textbf{deployment-friendly}, and \textbf{effective} for multimodal semantic segmentation?} 

We answer \emph{yes} to this question, and show a solution from the \emph{prompt learning view}. The idea of prompt learning (also known as prompt tuning) has originated from natural language processing (NLP) (\eg,~\cite{li2021prefix,powerprompt}) and aims to transfer the knowledge from frozen language models to various downstream tasks by injecting textual prompts in an efficiency way. Motivated by this, researchers have recently applied prompt tuning to vision tasks (\eg,~\cite{VPT, vipt, adaptformer, e2vpt, promptsegmentation, CoOp, CoCoOp}) and exhibited promising results. Despite this, these existing prompting learning methods are \emph{not} applicable to complicated multimodal dense prediction tasks demanding effective feature fusion and adaption, which thus motivates us to seek a new prompt tuning approach for multimodal semantic segmentation.

Considering multimodal semantic segmentation requires to deal with both modality fusion and feature adaption in the frozen backbone, we introduce \emph{\textbf{DPLNet}}, a novel and simple but effective \emph{Dual-Prompt Learning Network} that is able to adapt a frozen pre-trained model for multimodal semantic segmentation (see Fig.~\ref{fig:comp} (b)), and thus avoids massive parameter update. Specifically, DPLNet consists of two simple but crucial prompt learning modules, \ie, a \emph{multimodal prompt generator} (MPG) and a \emph{multimodal feature adapter} (MFA). MPG aims to fuse the important auxiliary modality feature into the RGB feature to generate the complementary multimodal feature prompt. To leverage different semantic-level features, multiple MPGs are inserted from shadow to deep stages, connected in a progressive fashion, to generate multi-level multimodal feature prompts, which are injected into the frozen network to guide semantic segmentation. In order to adapt the frozen single-modal backbone for better multimodal feature extraction, MFA is applied in each stage by introducing a few learnable tokens, which interact with multimodal features via cross-attention for adaption to a specific task, significantly enhancing performance. Notice that, different from VPT~\cite{VPT} which directly prepends the learnable tokens to each encoder layer, MFA applies an independent cross-attention to adapt multimodal features and obtains better results. Please note that, since both MPG and MFA are lightweight, only a few minimal number of trainable parameters (equivalent to 4.4\% of the original model parameters) are introduced in DPLNet. Using a simple decoder, DPLNet achieves superior results. Fig.~\ref{fig:framework} illustrates the architecture of DPLNet.

Compared with existing multimodal methods for semantic segmentation, DPLNet has several attractive properties: \textbf{First}, it is training-efficient architecture because only a few parameters requires to be tuned. \textbf{Second}, it largely reduces the deployment-burden in real-world applications as it does not need to retain dual encoders when deployed. \textbf{Third}, our DPLNet is a unified paradigm for multimodal semantic segmentation, and avoids complex task-specific model design.

To validate the effectiveness of DPLNet, we conduct extensive experiments on four challenging datasets, including NYUD-v2~\cite{nyuv2} and SUN-RGBD~\cite{sunrgbd} for RGB-D semantic segmentation and MFNet~\cite{mfnet} and PST900~\cite{pst900} for RGB-T semantic segmentation. DPLNet achieves state-of-the-art performance on NYUD-v2, SUN-RGBD, and PST900 and comparable results on MFNet while being efficient in training, evidencing its efficacy. Furthermore, we show that our DPLNet is general and applicable to other multimodal tasks such as salient object detection and video semantic segmentation, as shown in our experiments.

In summary, our \textbf{contributions} are as follows:
\vspace{0.1em}
\begin{itemize}
\setlength{\itemsep}{2pt}
\setlength{\parsep}{2pt}
\setlength{\parskip}{2pt}

      \item[$\spadesuit$] \emph{We propose a novel and simple yet effective dual-prompt learning paradigm, dubbed DPLNet, for training-efficient multimodal semantic segmentation.
      }
      
      \item[$\heartsuit$] \emph{We propose a multimodal prompt generator (MPG) module to fuse different modalities in a compact manner, eliminating complex multimodal fusion in previous methods.}

      \item[$\clubsuit$] \emph{We introduce a multimodal feature adapter (MFA) module to adapt the frozen pre-trained backbone for better multimodal feature extraction, greatly boosting performance.}
      
      \item[$\diamondsuit$] \emph{In extensive experiments on four multimodal benchmarks, DPLNet achieves state-of-the-art results or is on par with other methods while satisfying parameter efficiency, validating its effectiveness. Moreover, we show that DPLNet is general and applicable to other tasks.}
\end{itemize}

\section{Related Work}
\label{sec:rel}

In this section, we briefly discuss the multimodal semantic segmentation and visual prompt learning that are relevant to this work and their differences with the proposed DPLNet.

\subsection{Multimodal Semantic Segmentation}

\noindent
\textbf{RGB-D semantic segmentation} aims at improving the performance by introducing the depth information which consists of abundant geometric properties. Significant progress has been made by recent methods \cite{acnet, SGNet, SA-gate, cen, shapeconv, frnet, pgdenet, tokenfusion, multimae, omnivore, CMX, CMNext}. For example, the approach of \cite{acnet} leverages an encoder-decoder architecture and proposes an effective fusion module to fuse the information between two modalities. The work of \cite{tokenfusion} dynamically fuses the modality features by leveraging a transformer architecture. The works of \cite{CMX,CMNext} employ a similar spirit to utilize the transformer as the backbone to extract features, and then fuse the different modality information to achieve RGB-D semantic segmentation, demonstrating promising results.

\vspace{0.3em}
\noindent
\textbf{RGB-T semantic segmentation} focuses on improving semantic segmentation in low-light conditions using RGB and the auxiliary thermal modalities. Similar to the RGB-D semantic segmentation, the RGB-T-based methods mainly add an auxiliary branch based on RGB-based models to extract thermal features and then fuse the multimodal features for segmentation. Many methods \cite{mfnet, RTFNet, abmdrnet, EGFNet, GMNet, EGFNetTits, MTANet, EAEFNet, CMX, CMNext} have been recently proposed in this field. 
The approach of~\cite{RTFNet} and its follow-up works~\cite{EAEFNet, EGFNetTits, abmdrnet} propose to develop a dual-branch encoder-decoder architecture with a simple fusion strategy for RGB-T semantic segmentation. The approach of \cite{EGFNet} applies the edge prior map to enhance boundary information for improvement. Interestingly, the methods in \cite{CMX,CMNext} for RGB-D semantic segmentation are also applicable to the task of RGB-T semantic segmentation, indicating the underlying relationship between the two tasks and the demand for a unified framework.

\textbf{\emph{Different}} from the above multimodal RGB-D/T methods for semantic segmentation that fully fine-tune a dual-branch encoder-decoder architecture, our DPLNet aims at learning a parameter-efficient paradigm with dual-prompt tuning. In addition, our DPLNet is a unified framework that is applicable to various multimodal semantic segmentation tasks with promising performance.

\subsection{Visual Prompt Learning}

Prompt learning (also known as prompt tuning) has recently attracted extensive attention owing to its ability to decrease significantly the number of trainable parameters, which provides an efficient manner to leverage the pre-trained models. It originates from NLP~\cite{li2021prefix,powerprompt} and is applied to various vision tasks~\cite{VPT,vipt, adaptformer,e2vpt, promptsegmentation,yang2022prompting}. The work of~\cite{VPT} prepends a set of trainable parameters to adapt vision transformer (ViT)~\cite{dosovitskiy2020image} to various downstream visual recognition tasks, exhibiting remarkable performance. Different from~\cite{VPT}, the method of~\cite{adaptformer} proposes to insert a lightweight module into ViT and achieves superior results over the full fine-tuning models. The algorithm of~\cite{e2vpt} adds a few learnable parameters to the self-attention operation of the transformer model and applies pruning methods to reduce parameters, achieving effective and efficient prompt learning. Besides visual recognition, prompt tuning has been successfully applied in other visual tasks including tracking~\cite{vipt,yang2022prompting}, style transfer~\cite{sohn2023visual}, (RGB-based) semantic segmentation~\cite{promptsegmentation}, etc.

\textbf{\emph{Different}} from the aforementioned visual prompting tuning approaches, DPLNet is specifically focused on the task of multimodal dense prediction for RGB-D/T semantic segmentation, which to our knowledge has not been studied before. In addition to the difference in task, the proposed DPLNet introduces the novel dual-prompt learning, which differs from the above visual single-prompt tuning methods and displays superior performance for multimodal semantic segmentation, as can be seen in our experiments.

\section{The Proposed Approach}

\textbf{Overview.} In this paper, we introduce a simple yet effective approach, DPLNet, which enables tuning a RGB-based pre-trained model for multimodal (MM) semantic segmentation in a \emph{parameter-efficient} way. The key of DPLNet comprises two prompt tuning modules, including the MPG and MFA. As in Fig.~\ref{fig:framework}, MPG aims at fusing the features from multiple modalities and returns a multimodal feature prompt to the frozen model, while MFA works for adapting the prompted feature to the specific task for better performance. Because the main backbone is frozen, only a few additional trainable parameters are added, leading to high efficiency in training.

\subsection{Preliminaries}
\label{pre}

Before elaborating on the proposed method, we first present the necessary preliminaries for the used RGB-based pre-trained model and the problem definition for our task.

\textbf{Pre-trained RGB Model.} We focus on tuning the vision Transformer~\cite{segformer} (called $\textbf{T}_\text{RGB}$), which has been pre-trained with a large amount of data for RGB segmentation, for MM semantic segmentation. $\textbf{T}_\text{RGB}$ has two major components, including an encoder for feature extraction and a head for the specific task. The encoder consists of a stack of blocks, with each containing a set of self-attention layers. Specifically, given an RGB image $I_\text{RGB}$, $\textbf{T}_\text{RGB}$ first projects it into the token embedding $Z^0=\mathtt{Emb}(I_\text{RGB})$. Then it extracts features via $N$ encoder blocks $B_{i}(\cdot)$ ($i$=1,$\cdots$,$N$), followed by a decoder $\mathtt{Dec}(\cdots)$ for generating result $Y_\text{RGB}$, as follows,
\begin{equation*}\small
	\begin{split}
		Z^{i}&=B_{i}(Z^{i-1}) \;\;\; i=1,2,\cdots,N \\
		Y_\text{RGB} &= \mathtt{Dec}(Z^{1}, Z^{2}, \cdots, Z^{N}) 
	\end{split} \tag{1}
\end{equation*}
where $Z^i$ ($1\le i\le N$) is the output from encoder block $B_i$, and fed to $B_{i+1}$. Each encoder block $B_{i}$ is formed by a set of $K_i$ self-attention layers, with each defined as follows,
\begin{equation*}
	\mathtt{SAL}_{k}(z)=\mathtt{FFN}(\mathtt{MSA}(z)) \tag{2}
\end{equation*}
where $\mathtt{SAL}_{k}(\cdot)$ ($1\le k\le K_i$) denotes the $k^{\text{th}}$ self-attention layer, containing an efficient multi-head self-attention module $\mathtt{MSA}(\cdot)$ and a feed-forward network $\mathtt{FFN}(\cdot)$. We omit the layer normalization in self-attention layer for simplicity.

\textbf{Problem Definition.} We aim to adapt a pre-trained $\textbf{T}_\text{RGB}$ to learn multimodal semantic segmentation model $\textbf{T}_\text{RGB-X}$, which receives two images $I_\text{RGB}$ and $I_\text{X}$ from RGB and auxiliary X (X can be Depth or Thermal) modalities and output prediction $Y_\text{RGB-X}$, while keeping encoder of $\textbf{T}_\text{RGB}$ frozen.

\subsection{DPLNet for MM Semantic Segmentation}
\label{dplnet}

We propose DPLNet to adapt $\textbf{T}_\text{RGB}$ for efficient learning of $\textbf{T}_\text{RGB-X}$ for multimodal semantic segmentation. The core of DPLNet consists of two prompt-learning modules, including MPG and MFA. MPG aims to generate the multimodal RGB-X prompt, while MFA works on adapting the frozen encoder of $\textbf{T}_\text{RGB}$ for better MM semantic segmentation.

Specifically, given a pair of RGB-X input images $I_{RGB}$ and $I_{X}$, we first employ a patch embedding layer to obtain RGB and X token embeddings $Z_{RGB}^{0}$ and $Z_{X}^{0}$ via
\begin{equation*}\label{token}
		Z_{RGB}^0=\mathtt{Emb}_{RGB}(I_\text{RGB}) \;\;\;\;\;\; Z_{X}^0=\mathtt{Emb}_{Z}(I_\text{X})  \tag{3}
\end{equation*}
where $\mathtt{Emb}_{RGB}(\cdot)$ is the frozen patch embedding layer from $\textbf{T}_\text{RGB}$, and $\mathtt{Emb}_{Z}(\cdot)$ is the learnable patch embedding layer.

After that, a set of MPGs is employed in DPLNet, as shown in Fig.~\ref{fig:framework}, to learn a cascaded multimodal prompt, which is added to the original RGB flow with residual connection for the encoder block. In particular, for the $i^{\text{th}}$ encoder $B_i$, its input feature $Z^{i-1}$ is obtained by
\begin{equation*}
Z^{i-1} = Z_{RGB}^{i-1} + P^{i} \;\;\; i=1,2,\cdots,N  \tag{4}
\end{equation*}
where $ Z_{RGB}^{i-1}$ represents the output feature of the $(i-1)^{\text{th}}$ encoder $B_{i-1}$ (here $i>1$) and $P^{i}$ is the multimodal prompt generated from the $i^{\text{th}}$ MPG as follows,
\begin{equation*}
P^{i} = \mathtt{MPG}(Z_{RGB}^{i-1}, P^{i-1}) \tag{5}
\end{equation*}
where $\mathtt{MPG}(\cdot,\cdot)$ denotes our proposed MPG as explained later, and $P^{i-1}$ is the multimodal prompt generated from the last MPG, where $P^{0}=Z_X^0$.

With the multimodal prompt $Z^{i-1}$, we send it to $B_i$ for further feature learning. However, because $B_i$ is pre-trained only with RGB modality and frozen in DPLNet, it may not be suitable for multimodal feature learning. To address this issue, we introduce MFA to adapt $B_{i}$ for multimodal feature learning. The key idea is to leverage a small set of learnable prompts to generate the query-adaptive prompt and insert it into each self-attention layer of $B_i$. Specifically, for the $k^{\text{th}}$ self-attention layer $\mathtt{SAL}_{k}$ of $B_i$, a set of learnable prompt tokens $H_{k}^{i}$ are used to generate the adaption prompt via
\begin{equation*}
A_k^{i} = \mathtt{MFA}(H_{k}^{i}, Q_k^{i})  \tag{6}
\end{equation*}
where $\mathtt{MFA}(\cdot,\cdot)$ represents our MFA as describe late, $Q_k^{i}=W_{q}{\mathtt{SAL}_{k-1}(z)}$ is the query generated by last self-attention layer, and $A_k^{i}$ is the generated adaption prompt. With $A_k^{i}$, we insert it into $\mathtt{SAL}_{k}$ with residual addition via
\begin{equation*}
	\mathtt{SAL}_{k}(z)=\mathtt{FFN}(\mathtt{MSA}(z)+A_k^{i})  \tag{7}
\end{equation*}
With the above equation, we can adapt $B_{i}$ to better multimodal feature learning by $Z_{RGB}^{i}=B_{i}(Z^{i-1})$, where $Z_{RGB}^i$ is the output feature after the encoder block $B_{i}$, and will be fed to the next encoder block. 

After the $N^{\text{th}}$ encoder block, all features $Z_{RGB}^{i}$ are sent to the decoder for segmentation as in $\mathbf{T}_{\text{RGB}}$ via 

\begin{equation*}\small
		Y_\text{RGB-X} = \mathtt{Dec}(Z_{{RGB}}^{1}, Z_{{RGB}}^{2}, \cdots, Z_{{RGB}}^{N}) 
 \tag{8}
\end{equation*}
Notice that, $Z_{\text{RGB}}^{i}$ ($1\le i\le N$) here contains information from both RGB and X modalities. Due to the domain gap between RGB and RGB-X tasks, the decoder used for RGB-X semantic segmentation in this work is set to be learnable. During training, we use a simple cross-entropy loss to update parameters in DPLNet. Fig.~\ref{fig:framework} illustrates our DPLNet.

\begin{figure}[t]
 \centering
 \includegraphics[width=1.0\linewidth]{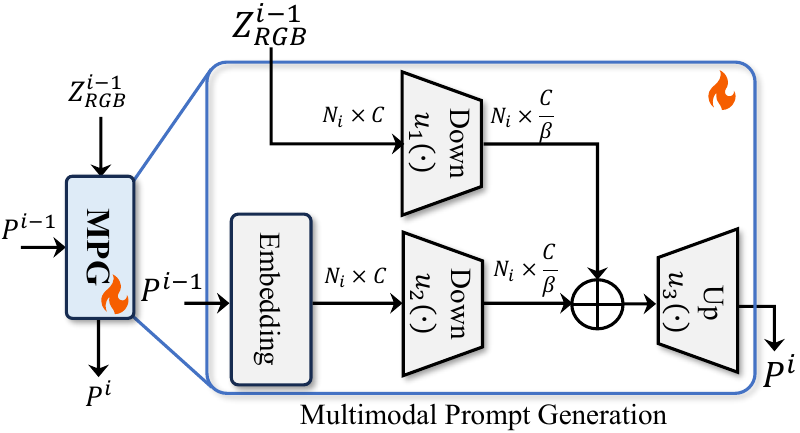}
 \caption{The structure of MPG, which aims to fuse RGB features with a multimodal prompt from the last stage for generating a new multimodal prompt.}
 \label{fig:MPG}
 \vspace{-2mm}
\end{figure}

\subsection{Multimodal Prompt Generation}
\label{mpg}
A key of our DPLNet is to generate appropriate multimodal prompts for guiding the network for effective MM semantic segmenting. To this end, a multimodal prompt generation (MPG) is proposed by fusing multimodal information in a compact way, which is thus more parameter-efficient than existing heavy fusion strategies. Fig.~\ref{fig:MPG} shows the architecture of MPG.
More concretely, given the RGB features $Z_{RGB}^{i-1}\in R^{{N_i}\times{C}}$ generated from encoder $B_{i-1}$ and the multimodal prompt $P^{i-1} \in R^{{N_i}\times{C}}$ generated from last MPG module, where $N_i$ denotes the token number and $C$ denotes the token dimension, we follow the work of \cite{segformer} to adopt a patch embedding strategy (which is implemented using a light ${3{\times}3}$ convolutional layer) to obtain more robust multimodal prompt information:
\begin{equation*}
	\hat{P}^{i-1} =\mathtt{Emb_i}(P^{i-1}) \;\;\;i=2,3,\cdots,N \tag{9} 
\end{equation*}
where $\mathtt{Emb_i}(\cdot)$ denotes the learnable patch embedding layer. We omit the patch embedding layer in the first MPG module because the $I_{X}$ is tokenized by $\mathtt{Emb}_{Z}(\cdot)$ in Eq.~\ref{token}.

Inspired by~\cite{adaptformer}, we propose to downsample features from different modalities in channel dim for fusion and then upsample the fused feature back to the original dimension for feature adaption. Mathematically, this process can be formulated as follows,
\begin{equation*}
	P^{i}={u_3}(u_1(Z_{RGB}^{i-1})+ u_2(\hat{P}^{i-1})) \;\;\;i=1,2,\cdots,N \tag{10}
\end{equation*}
where $u_{1}(\cdot)$ and $u_{2}(\cdot)$ are two projection layers to reduce feature dimension, while $u_{3}(\cdot)$ works to increase the feature dimension. All the parameters are learnable in MPG module and the reducing factor $\beta$ is set to 4.

\begin{figure}[t]
 \centering
 \includegraphics[width=1.0\linewidth]{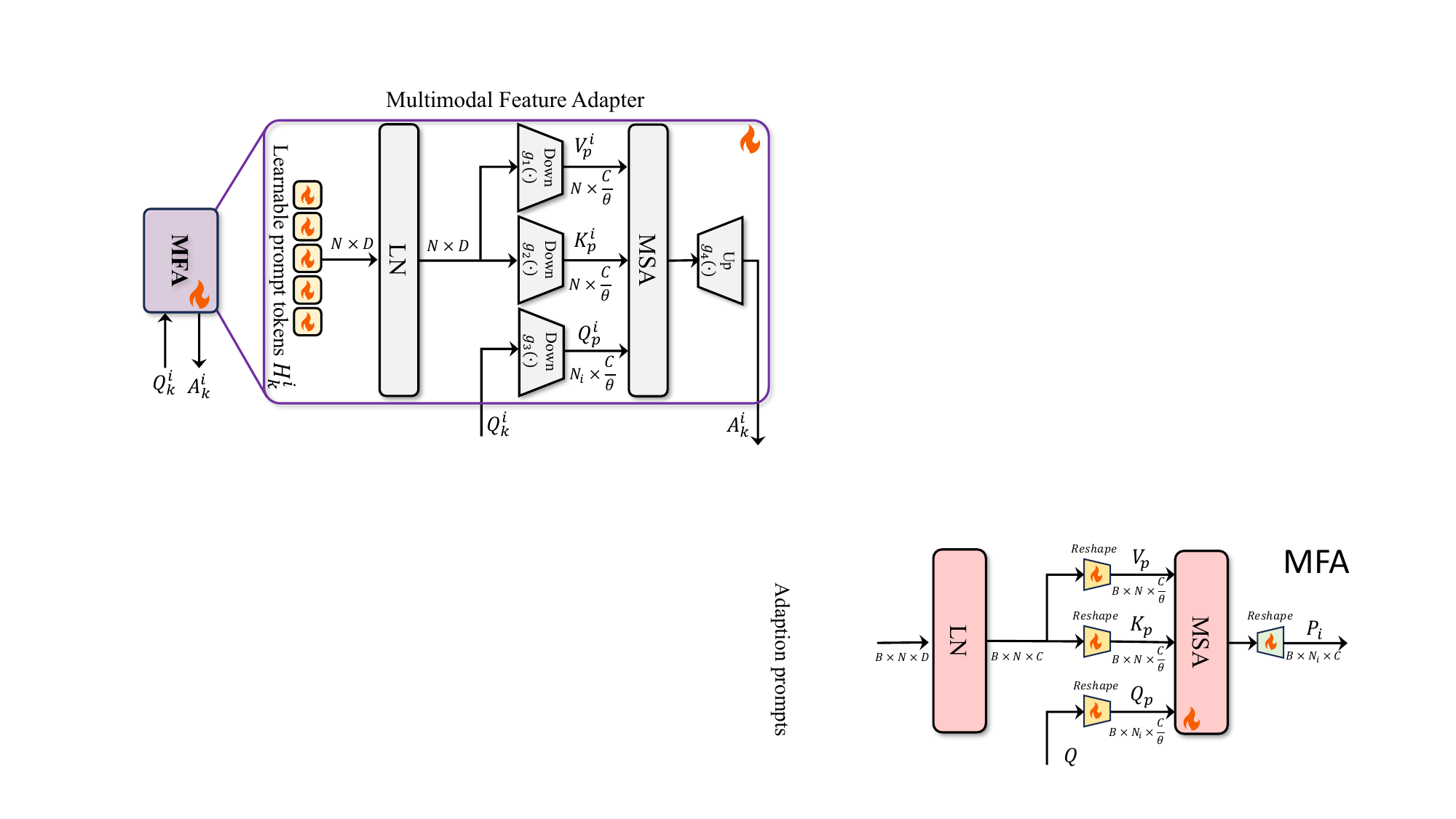}
 \caption{The structure of MFA, which is leveraged to adapt the query features in frozen backbone with a set of learnable tokens.}
 \label{fig:MFA}
 \vspace{-2mm}
\end{figure}

\subsection{Multimodal Feature Adapter}
\label{mfa}
In our DPLNet, the generated multimodal prompt is sent to the pre-trained RGB Transformer block for feature learning. Due to the domain gap between different tasks, this may not be optimal. To deal with this, we propose a multimodal feature adapter (MFA) that inserts a set of learnable prompt tokens to improve feature learning for pre-trained RGB Transformer blocks. It is worth noticing that, compared to the recent visual prompt tuning (VPT)~\cite{VPT} in token space for visual recognition, the proposed MFA differs by adapting queries only for dense prediction, which is more effective. More specifically, as shown in Fig.~\ref{fig:MFA}, we generate adaption prompt $A_k^{i} \in R^{B\times{N_i}\times{C}}$ based on original self-attention query $Q_{k}^{i} \in R^{B\times{N_i}\times{C}}$, which can learn query-adaptive information from the learnable prompt tokens $H_{k}^{i} \in R^{B\times{N}\times{D}}$, where $N$ and $D$ are the number and dimension of learnable prompt tokens, respectively. 

Drawing inspiration from~\cite{LORA} which shows that the pre-trained model resides on a low intrinsic dimension, we obtain the prompt query {$Q_p^{i}$} by projecting self-attention query $Q_{k}^{i}$ via a linear projection layer to reduce the redundant features and parameters through the reducing factor $\theta$ via
\begin{equation*}
Q_p^{i}=g_3(Q_{k}^{i})\;\;\;i=1,2,\cdots,N\tag{11}
\end{equation*}
Accordingly, we also reduce the dimension of {$K_p^{i}$} and {$V_p^{i}$} by linear projection layer through the reducing factor $\theta$ after a normalization layer as follows,
\begin{equation*}
	V_p^{i}=g_1(\mathtt{LN}(H_{k}^{i})) \tag{12}
\end{equation*}
\begin{equation*}
	K_p^{i}=g_2(\mathtt{LN}(H_{k}^{i})) \tag{13}
\end{equation*}
where $g_1(\cdot)$, $g_2(\cdot)$ and $g_3(\cdot)$ are liner projection layers, $\mathtt{LN}(\cdot)$ is the normalization layer and the reducing factor $\theta$ is set to 32. After that, we expand the dimension into original embedding space using $g_4(\cdot)$ after the MSA layer to generate the adaption prompt $A_{k}^{i}$ as follows,
\begin{equation*}
A_{k}^{i} = g_{4}(\mathtt{MSA}(Q_p^{i}, K_p^{i}, V_p^{i})) \tag{14}
\end{equation*}

\section{Experiments}
\label{sec:Experiments}

\noindent
\textbf{Datasets.} To validate the effectiveness of our approach, we conduct extensive experiments on four datasets, including two RGB-D datasets NYUDv2~\cite{nyuv2} and SUN RDB-D~\cite{sunrgbd} and two RGB-T datasets MFNet~\cite{mfnet} and PST900~\cite{pst900}.

\begin{table*}[h]
\LARGE
\centering
\setlength{\tabcolsep}{5pt}
\renewcommand{\arraystretch}{1.1}
\scalebox{0.415}{
\begin{tabular}{c c c c c c c c c c c c c c}\\
\hline
 &ACNet &SGNet &SA-Gate &PGDENet &TokenFusion-B3 &MultiMAE &Omnivore-B &CMX-B5 &CMXNeXt  &DFormer-L &DPLNet &DPLNet\\ 
 &-SS \cite{acnet}  &-MS \cite{SGNet} &-MS \cite{SA-gate} &-SS \cite{pgdenet} &-SS \cite{tokenfusion} &-SS \cite{multimae} &-SS \cite{omnivore} &-MS \cite{CMX} &-MS \cite{CMNext} &-MS \cite{DFormer} &-SS (Ours) &-MS (Ours)\\
\hline\hline
\multicolumn{1}{c}{Backbone} &ResNet-50 &ResNet-101 &ResNet-101 &ResNet-34 &MiT-B3 &ViT-B &Swin-B   &MiT-B5  &MiT-B4 &DFormer-L &MiT-B5 &MiT-B5 
\\
\multicolumn{1}{c}{Params (M)} &116.6 &64.7  &110.9 &100.7 &45.9  &95.2 &95.7 &181.1 &119.6 &39.0 &\textbf{7.15} &\textbf{7.15}
\\
\multicolumn{1}{c}{mIoU (\%)} &48.3 &51.1 &52.4 &53.7  &54.2 &56.0 &54.0 &56.9 &56.9 &57.2 &\textbf{\color{blue}58.3} &\textbf{\color{red}59.3}
\\
\hline
\end{tabular}
}
\caption{\centering RGB-D semantic segmentation on NYUDv2. ``SS'' and ``MS'' denote single- or multi-scale for the test of RGB-D segmentation. ``Params'' represents learnable parameters in all tables. The best two results are highlighted in \textbf{\color{red}{red}} and \textbf{\color{blue}{blue}} in all the comparison tables.}
\label{tb:1}
\vspace{-1mm}
\end{table*}

\begin{table*}[h]
\LARGE
\centering
\setlength{\tabcolsep}{5pt}
\renewcommand{\arraystretch}{1.1}
\scalebox{0.415}{
\begin{tabular}{c c c c c c c c  c c c c c c}\\
\hline
 &ACNet &SGNet &SA-Gate &PGDENet &TokenFusion-B3 &CMX-B4 &CMX-B5 &CMXNeXt &DFormer-B  &DFormer-L &DPLNet &DPLNet\\ 
 &-SS \cite{acnet}  &-MS \cite{SGNet} &-MS \cite{SA-gate} &-SS \cite{pgdenet} &-SS \cite{tokenfusion} &-MS \cite{CMX} &-MS \cite{CMX} &-MS \cite{CMNext} &-MS \cite{DFormer} &-MS \cite{DFormer} &-SS (Ours)  &-MS (Ours)\\
\hline\hline
\multicolumn{1}{c}{Backbone} &ResNet-50 &ResNet-101 &ResNet-101 &ResNet-34 &MiT-B3 &MiT-B4 &MiT-B5   &MiT-B4  &DFormer-B &DFormer-L &MiT-B5 &MiT-B5 
\\
\multicolumn{1}{c}{Params (M)} &116.6 &64.7  &110.9 &100.7 &45.9  &139.9 &181.1 &119.6 &29.5 &39.0 &\textbf{7.15} &\textbf{7.15}
\\
\multicolumn{1}{c}{mIoU (\%)} &48.1 &48.6 &49.4 &51.0  &51.0† &52.1 &52.4 &51.9† &51.2 &\textbf{\color{blue}52.5} &52.1 &\textbf{\color{red}52.8}
\\
\hline
\end{tabular}
}
\caption{\centering RGB-D semantic segmentation on SUN RGB-D. † indicates that we follow the results from DFormer \cite{DFormer}.}
\label{tb:2}
\vspace{-1mm}
\end{table*}

\begin{table*}[!t]
\LARGE
\centering
\setlength{\tabcolsep}{7.5pt}
\renewcommand{\arraystretch}{1.1}
\scalebox{0.415}{
\begin{tabular}{c c c c c c c c c c c c c}\\
\hline
 &MFNet &RTFNet &FuseSeg-161 &ABMDRNet &EGFNet &MTANet &GEBNet &GMNet &CMX-B2 &CMX-B4 &CMNeXt &DPLNet \\ 
 &\cite{mfnet}  &\cite{RTFNet} &\cite{fuseseg} &\cite{abmdrnet} &\cite{EGFNet} &\cite{MTANet} &\cite{GEBNet} &\cite{GMNet} &\cite{CMX} &\cite{CMX} &\cite{CMNext} &(Ours)\\
\hline\hline
\multicolumn{1}{c}{Backbone} &- &ResNet-152 &DenseNet-161 &ResNet-50 &ResNet-152 &ResNet-152 &ConvNeXt-S  &ResNet-50  &MiT-B2 &MiT-B4 &MiT-B4 &MiT-B5 
\\
\multicolumn{1}{c}{Params (M)} &8.4 &337.1  &141.5 &- &201.3  &121.9 &- &149.8 &66.6 &139.9 &119.6 &\textbf{7.15}
\\
\multicolumn{1}{c}{mIoU (\%)} &39.7 &53.2 &54.5 &54.8  &54.8 &56.1 &56.2 &57.3 &58.2 &\textbf{\color{blue}59.7} &\textbf{\color{red}59.9} &59.3 
\\
\hline
\end{tabular}
}
\caption{\centering RGB-T semantic segmentation results on MFNet benchmark.}
\label{tb:3}
\vspace{-1mm}
\end{table*}

\begin{table*}[!t]
\LARGE
\centering
\setlength{\tabcolsep}{7pt}
\renewcommand{\arraystretch}{1.1}
\scalebox{0.49}{
\begin{tabular}{c c c c c c c c c c}\\
\hline
 &RTFNet &PSTNet &MTANet &GMNet  &EGFNet &EGFNet-ConvNext  &GEBNet &CACFNet &DPLNet \\ 
 &\cite{RTFNet}  &\cite{pst900} &\cite{MTANet} &\cite{GMNet} &\cite{EGFNet} &\cite{EGFNetTits} &\cite{GEBNet} &\cite{Cacfnet} &(Ours)\\
\hline\hline
\multicolumn{1}{c}{Backbone} &ResNet-152 &ResNet-18 &ResNet-152  &ResNet-50 &ResNet-152 &ConvNeXt-B &ConvNeXt-S  &ConvNeXt-B  &MiT-B5
\\
\multicolumn{1}{c}{Params (M)} &337.1 &105.8  &121.9 &149.8 &201.3  &- &- &198.6  &\textbf{7.15}
\\
\multicolumn{1}{c}{mIoU (\%)} &60.5 &68.4 &78.6 &84.1  &78.5 &85.4 &81.2 &\textbf{\color{blue}86.6} &\textbf{\color{red}86.7} 
\\
\hline
\end{tabular}
}
\caption{\centering RGB-T semantic segmentation results on PST900 benchmark.}
\label{tb:4}
\vspace{-1mm}
\end{table*}

\begin{table}[!t]
\Large
\centering
\setlength{\tabcolsep}{10pt}
\renewcommand{\arraystretch}{1.05}
\scalebox{0.7}{
\begin{tabular}{c c c c}
\hline
\multicolumn{2}{c}{{Methods}} &{Params (M)} &{mIoU (\%)}
\\
\hline\hline
\multicolumn{2}{c}{w/o MPG} &6.86  &57.4
\\
\multicolumn{2}{c}{w/o MFA} &5.49  &57.4
\\
\multicolumn{2}{c}{Frozen decoder} &3.88  &55.1
\\
\multicolumn{2}{c}{Fully fine-tuning} &88.58  &58.1
\\
\multicolumn{2}{c}{DPLNet (Ours)} &7.15 &58.3
\\
\hline
\end{tabular}
}
\caption{\centering Ablation of key modules in DPLNet.}
\label{tb:5}
\vspace{-4mm}
\end{table}

NYUDv2~\cite{nyuv2} consists of 1,449 RGB-D samples, among which 795 are used for training and the rest for testing. It has 41 categories including a background class. SUN RGB-D~\cite{sunrgbd} comprises 10,335 labeled RGB-D images which are divided into 5,285 and 5,050 RGB-D pairs for training and testing, respectively. MFNet~\cite{mfnet} has 1,569 RGB-T images including 820 daytime image pairs and 749 nighttime image pairs from eight foreground classes and one background class. PST900~\cite{pst900} contains 894 aligned pairs of RGB and thermal images with a spatial resolution of 720 × 1280 pixels. These images are from five categories. We follow the training/test split as in~\cite{pst900} for our experiment.

\vspace{0.3em}
\noindent
\textbf{Evaluation Metric.} Following existing methods, we report the popular Intersection over Union (IoU) of each method when conducting a comparison.

\vspace{0.3em}
\noindent{\textbf{Implementation.}}
We conducted all experiments on a single NVIDIA A6000 GPU with PyTorch 1.12. For NYUDv2 \cite{nyuv2}, the learning rate is set as 4e{-2} and the weight decay is 5e{-4}. We train DPLNet for 400 epochs with a batch size of 20. For SUN RGB-D \cite{sunrgbd}, we set the learning rate as 1e{-2} and the weight decay is 5e{-4}. We train our network on this benchmark for 300 epochs and the batch size is also set to 20. Following many existing RGB-D semantic segmentation works (\eg,~\cite{DFormer}), we adopt the multi-scale (MS) flip inference strategies with scales \{0.5, 0.75, 1, 1.25, 1.5\}, but also report the result of our single-scale (SS) version.
For MFNet \cite{mfnet}, the learning rate is 5e{-3} with a weight decay of 5e{-4} and a batch size of 16. We train our DPLNet for 700 epochs.
For PST900 \cite{pst900}, the learning rate is set to 1e{-3} with a weight decay of 5e{-4}. We train our network for 400 epochs and the batch size is set to 6. During training for all the experiments, we utilize the same network architecture and it adds only a few learnable parameters of 7.15M, with 3.88M for adapting the backbone which is only 4.4\% of the original backbone, and 3.27M for the decoder.

\begin{table*}[h]
    \begin{subtable}[h]{0.33\textwidth}
        \centering
        \caption{\centering Ablation studies for the position of MPG module (Bottom to top).}
        \scalebox{0.875}{
        \begin{tabular}{c c c c} 
        \hline
        \multicolumn{2}{c}{{Position}} &{Params} &{mIoU (\%)}
        \\
        \hline\hline
        \multicolumn{2}{c}{Stage 1}  &{6.86}  &57.6 \\
        \multicolumn{2}{c}{Stage 1$\,\to\,$2}  &{6.87}  &57.8 \\
        \multicolumn{2}{c}{Stage 1$\,\to\,$3}  &{6.95}  &57.6 \\
         \multicolumn{2}{c}{Stage 1$\,\to\,$4 (Ours)}  &{7.15}  &58.3
        \\
        \hline
       \end{tabular}
       \label{tb:6(a)}
       }
    \end{subtable}
    \hfill
    \begin{subtable}[h]{0.33\textwidth}
        \centering
        \caption{\centering Ablation studies for the position of MPG module (Top to bottom).}
        \scalebox{0.875}{
        \begin{tabular}{c c c c} \hline
        \multicolumn{2}{c}{{Position}} &{Params} &{mIoU (\%)}
        \\
        \hline\hline
        \multicolumn{2}{c}{Stage 4}  &{7.05}  &57.3 \\
        \multicolumn{2}{c}{Stage 4$\,\to\,$3}  &{7.13}  &57.2 \\
        \multicolumn{2}{c}{Stage 4$\,\to\,$2}  &{7.14}  &57.8 \\
        \multicolumn{2}{c}{Stage 4$\,\to\,$1 (Ours)}  &{7.15}  &58.3
        \\
        \hline
        \end{tabular}
        \label{tb:6(b)}}
    \end{subtable}
    \begin{subtable}[h]{0.33\textwidth}
        \centering
        \caption{\centering Ablation studies of learnable prompt length in MFA module.}
        \scalebox{0.875}{
        \begin{tabular}{c c c c} \hline
        \multicolumn{2}{c}{{Prompt length}} &{Params} &{mIoU (\%)}
        \\
        \hline\hline
        \multicolumn{2}{c}{10} &7.1456  &57.8
        \\
        \multicolumn{2}{c}{20} &7.1460  &57.5
        \\
        \multicolumn{2}{c}{30 (Ours)} &7.1463  &58.3
        \\
        \multicolumn{2}{c}{40} &7.1466 &57.7
        \\
        \hline
        \end{tabular}
        \label{tb:6(c)}}
    \end{subtable}
    \begin{subtable}[h]{0.33\textwidth}
        \centering
        \caption{\centering Ablation studies of learnable prompt dimension in MFA module.}
        \scalebox{0.875}{
        \begin{tabular}{c c c c} \hline
        \multicolumn{2}{c}{{Prompt dimension}} &{Params} &{mIoU (\%)}
        \\
        \hline\hline
        \multicolumn{2}{c}{8} &6.41 &57.7
        \\
        \multicolumn{2}{c}{16} &6.65 &57.9
        \\
        \multicolumn{2}{c}{32 (Ours)} &7.15 &58.3
        \\
        \multicolumn{2}{c}{64} &8.13 &57.8
        \\
        \hline
        \end{tabular}
        \label{tb:6(d)}}
    \end{subtable}
    \begin{subtable}[h]{0.33\textwidth}
        \centering
        \caption{\centering Ablation studies for the position of MFA module (Bottom to top).}
        \scalebox{0.875}{
        \begin{tabular}{c c c c} \hline
        \multicolumn{2}{c}{{Position}} &{Params} &{mIoU (\%)}
        \\
        \hline\hline
        \multicolumn{2}{c}{Stage 1}  &{5.51}  &57.4 \\
        \multicolumn{2}{c}{Stage 1$\,\to\,$2}  &{5.57}  &57.7\\
        \multicolumn{2}{c}{Stage 1$\,\to\,$3}  &{6.94}  &57.8
         \\
        \multicolumn{2}{c}{Stage 1$\,\to\,$4 (Ours)}  &{7.15}  &58.3
        \\
        \hline
        \end{tabular}
        \label{tb:6(e)}
        }
    \end{subtable}
    \begin{subtable}[h]{0.33\textwidth}
        \centering
        \caption{\centering Ablation studies for the position of MFA module (Top to bottom).}
        \scalebox{0.875}{
        \begin{tabular}{c c c c} \hline
        \multicolumn{2}{c}{{Position}} &{Params} &{mIoU (\%)}
        \\
        \hline\hline
        \multicolumn{2}{c}{Stage 4}  &{5.69}  &57.7 \\
        \multicolumn{2}{c}{Stage 4$\,\to\,$3}  &{7.07}  &57.6 \\
        \multicolumn{2}{c}{Stage 4$\,\to\,$2}  &{7.13}  &57.7 \\
        \multicolumn{2}{c}{Stage 4$\,\to\,$1 (Ours)}  &{7.15}  &58.3
        \\
        \hline
        \end{tabular}
        \label{tb:6(f)}
        }
    \end{subtable}
     \caption{Ablation studies on different settings for the proposed DPLNet using single-scale. Note that, {$i \to j$} in (a), (b), (e), and (f) indicates the which layer the MPG module is inserted into. The parameters in each table are measured by ``M''.}
     \label{tb:6}
\end{table*}

\subsection{State-of-the-art Comparison}
\label{4.2}
\noindent{\textbf{DPLNet for RGB-D semantic segmentation.}} For RGB-D semantic segmentation, we evaluate our method on two popular, benchmarks including NYUDv2 \cite{nyuv2} and SUN RGB-D \cite{sunrgbd}, and compare it with many recent state-of-the-art (SOTA) RGB-D semantic segmentation approaches. As demonstrated in Tab.~\ref{tb:1}, our DPLNet, even with single-scale inference, achieves promising results with 0.583 mIoU on NYUDv2. When applying a multi-scale inference strategy, DPLNet further improves the result to 0.593 mIoU, which significantly outperforms other models. In addition, DPLNet only contains 7.15M learnable parameters, which is more efficient than other approaches in training. Likewise, on the more challenging SUN RGB-D as shown in Tab.~\ref{tb:2}, our DPLNet with multi-scale inference achieves the best performance with 0.528 mIOU yet has much less trainable parameters compared to existing approaches. All these show the effectiveness of our DPLNet.

\vspace{0.3em}
\noindent{\textbf{DPLNet for RGB-T semantic segmentation.}} For RGB-T semantic segmentation, we conduct experiments on widely used MFNet~\cite{mfnet} and PST900~\cite{pst900}. Note that, for RGB-T semantic segmentation, all methods are evaluated in the setting of single-scale inference. Tab.~\ref{tb:3} demonstrates and compares our DPLNet with 11 SOTA RGB-T semantic segmentation models on MFNet. CMNexT shows the best result with mIoU of 0.599. Compared to CMNext, DPLNet achieves comparable performance with 0.593 mIoU, while significantly reducing the number of trainable parameters by 16$\times$ (7.15M \emph{v.s.} 119.6M), which shows the effectiveness of our approach. On PST900 as shown in Tab.~\ref{tb:4}, our DPLNet is compared to 8 recent models. From Tab.~\ref{tb:4}, we can clearly observe that DPLNet achieves the best performance with mIoU of 0.867, outperforming the second best CACFNet with 0.866 mIoU yet comprising much fewer parameters, evidencing the efficacy of the proposed method.

\begin{table}[!t]
\Large
\centering
\setlength{\tabcolsep}{10pt}
\renewcommand{\arraystretch}{1.05}
\scalebox{0.7}{
\begin{tabular}{c c c c}
\hline
\multicolumn{2}{c}{{Methods}} &{Params (M)} &{mIoU (\%)}
\\
\hline\hline
\multicolumn{2}{c}{CCNet \cite{CCNet}} &-  &51.7
\\
\multicolumn{2}{c}{OCRNet \cite{OCRNet}} &-  &52.4
\\
\multicolumn{2}{c}{STM \cite{STM}} &-  &52.5
\\
\multicolumn{2}{c}{LMANet \cite{LMANet}} &-  &52.7
\\
\multicolumn{2}{c}{MFNet \cite{mfnet}} &8.4 &51.6
\\
\multicolumn{2}{c}{RTFNet \cite{RTFNet}} &337.1 &52.8
\\
\multicolumn{2}{c}{EGFNet \cite{EGFNet}} &201.3 &53.4
\\
\multicolumn{2}{c}{MVNet \cite{MVNet}} &88.4 &\textbf{\color{blue}54.5} 
\\
\multicolumn{2}{c}{DPLNet (Ours)} &7.15 &\textbf{\color{red}57.9} 
\\
\hline
\end{tabular}
}
\caption{\centering RGB-T video semantic segmentation results on MVSeg.}
\label{tb:7}
\vspace{-4mm}
\end{table}

\begin{table*}[t]\small
\centering
\label{table1}
\setlength{\tabcolsep}{3pt}
\renewcommand{\arraystretch}{1.1}
\begin{tabular}{@{}c c c c c c c |c c c c |c c c c |c c c c |c c c c@{}}
\hline
\multicolumn{3}{c}{\multirow{2}{*}{Model}}  &\multicolumn{4}{c}{NJU2K \cite{(NJU2K}} &\multicolumn{4}{c}{NLPR \cite{NLPR}} &\multicolumn{4}{c}{DES \cite{DES}} &\multicolumn{4}{c}{SIP \cite{SIP}} &\multicolumn{4}{c}{LFSD \cite{LFSD}} \\
\cline{4-23}

& &   &$\textit{S}\uparrow$ &$\textit{E}\uparrow$ &$\textit{F}\uparrow$ &$\textit{M}\downarrow$ &$\textit{S}\uparrow$ &$\textit{E}\uparrow$ &$\textit{F}\uparrow$ &$\textit{M}\downarrow$ &$\textit{S}\uparrow$ &$\textit{E}\uparrow$ &$\textit{F}\uparrow$ &$\textit{M}\downarrow$ &$\textit{S}\uparrow$ &$\textit{E}\uparrow$ &$\textit{F}\uparrow$ &$\textit{M}\downarrow$ &$\textit{S}\uparrow$ &$\textit{E}\uparrow$ &$\textit{F}\uparrow$ &$\textit{M}\downarrow$\\
\hline\hline

\multicolumn{3}{c}{CMWNet \cite{CMWNet}}	&.903 &.912	&.880	&.046	&.917	&{.951}	&.872	&.029 &\textbf{\color{blue}.933}	&{.967}	&{.899}	&{.022} &.868	&.907	&{.851}	&.062 &\textbf{\color{red}.876}	&{.891}	&\textbf{\color{red}.871}	&\textbf{\color{blue}.067}
\\ 
\multicolumn{3}{c}{cmWS \cite{cmMS}}		&.900 &.914	&.886	&.044	&.915	&.945	&.870	&.027 &-	&-	&-	&- &-	&-	&-	&- &-	&-	&-	&-
\\
\multicolumn{3}{c}{SSF \cite{SSF}}		&.898 &.912	&.885	&.043	&.913	&.949	&{.875}	&{.026} &.903	&.946	&.882	&.026 &-	&-	&-	&- &{.858}	&\textbf{\color{blue}.895}	&\textbf{\color{blue}.866}	&\textbf{\color{red}.066}
\\
\multicolumn{3}{c}{BBSNet \cite{BBSNet}}		&\textbf{\color{blue}.912} &{.919}	&{.893}	&{.040}	&\textbf{\color{blue}.920}	&.945	&.870	&.027 &.906	&.941	&.866	&.029 &{.871}	&{.909}	&.850	&{.057} &.843	&.879	&.830	&{.081}
\\
\multicolumn{3}{c}{LSNet \cite{LSNet}}	&{.911} &\textbf{\color{blue}.922}	&\textbf{\color{blue}.900}	&\textbf{\color{blue}.037}	&{.918}	&\textbf{\color{blue}.956}	&\textbf{\color{blue}.885}	&\textbf{\color{blue}.024} &{.925}	&\textbf{\color{blue}.970}	&\textbf{\color{blue}.910}	&\textbf{\color{blue}.020}  &\textbf{\color{blue}.886}	&\textbf{\color{blue}.927}	&\textbf{\color{blue}.884}	&\textbf{\color{blue}.048} &.833	&.873	&.852	&.084
\\
\multicolumn{3}{c}{DPLNet (Ours)}	&\textbf{\color{red}.920} &\textbf{\color{red}.944}	&\textbf{\color{red}.904}	&\textbf{\color{red}.035}	&\textbf{\color{red}.933}	&\textbf{\color{red}.962}	&\textbf{\color{red}.897}	&\textbf{\color{red}.020} &\textbf{\color{red}.940}	&\textbf{\color{red}.978}	&\textbf{\color{red}.921}	&\textbf{\color{red}.017} &\textbf{\color{red}.890}	&\textbf{\color{red}.932}	&\textbf{\color{red}.888}	&\textbf{\color{red}.045} &\textbf{\color{blue}.873}	&\textbf{\color{red}.909}	&{.864}	&\textbf{\color{blue}.067}
\\
\hline
\end{tabular}
\caption{\centering Results and comparison on RGB-D SOD benchmarks. ${\uparrow }/ {\downarrow}$ indicates that a larger/smaller value is better.}
\label{tb:8}
\end{table*}

\begin{table*}[t]\small
\centering
\label{table1}
\setlength{\tabcolsep}{9.3pt}
\renewcommand{\arraystretch}{1.05}
\begin{tabular}{ c c  c c c c |c c c c |c c c c}
\hline
\multicolumn{2}{c}{\multirow{2}{*}{Model}}  &\multicolumn{4}{c}{VT821 \cite{VT821}} &\multicolumn{4}{c}{VT1000 \cite{VT1000}} &\multicolumn{4}{c}{VT5000 \cite{VT5000}} \\
\cline{3-14}

&    &$\textit{S}\uparrow$ &$\textit{E}\uparrow$ &$\textit{F}\uparrow$ &$\textit{M}\downarrow$ &$\textit{S}\uparrow$ &$\textit{E}\uparrow$ &$\textit{F}\uparrow$ &$\textit{M}\downarrow$ &$\textit{S}\uparrow$ &$\textit{E}\uparrow$ &$\textit{F}\uparrow$ &$\textit{M}\downarrow$
\\
\hline\hline
\multicolumn{2}{c}{SGDL \cite{VT1000}}	 &.765	&.847	&.731	&.085	&.787	&.856	&.764	&.090	&.750	&.824	&.672	&.089
\\

\multicolumn{2}{c}{PoolNet \cite{PoolNet}}	 &.751	&.739	&.578	&.109	&.834	&.813	&.714	&.067	&.769	&.755	&.588	&.089
\\

\multicolumn{2}{c}{R3Net \cite{R3net}}	 &.786 &.809	&.660	&.073	&.842	&.859	&.761	&.055	&.757	&.790	&.615	&.083
\\

\multicolumn{2}{c}{CPD \cite{CPD}}	&.827	&.837	&.710	&.057	&.906	&.902	&.834	&.032	&.848	&.867	&.741	&.050
\\

\multicolumn{2}{c}{MMCI \cite{MMCI}}	 &.763	&.784	&.618	&.087	&.886	&.892	&.803	&.039	&.827	&.859	&.714	&.055
\\
\multicolumn{2}{c}{AFNet \cite{AFNet}}	 &.778	&.816	&.661	&.069	&.888	&.912	&.838	&.033	&.834	&.877	&.750	&.050
\\

\multicolumn{2}{c}{TANet \cite{TANet}} &.818	&.852	&.717	&.052	&.902	&.912	&.838	&.030	&.847	&.883	&.754	&.047
\\

\multicolumn{2}{c}{S2MA \cite{S2MA}}	 &.811	&.813	&.709	&.098	&{.918}	&.912	&.848	&{.029}	&.853	&.864	&.743	&.053
\\

\multicolumn{2}{c}{JLDCF \cite{JLDCF}}	&.839	&.830	&.726	&.076	&.912	&.899	&.829	&.030	&.861	&.860	&.739	&.050
\\

\multicolumn{2}{c}{FMCF \cite{FMCF}} &.760	&.796	&.640	&.080	&.873	&.899	&.823	&.037	&.814	&.864	&.734	&.055
\\

\multicolumn{2}{c}{ADF \cite{VT5000}}	 &.810	&.842	&.717	&.077	&.910	&.921	&.847	&.034	&.864	&{.891}	&.778	&.048
\\

\multicolumn{2}{c}{MIDD \cite{MIDD}}	&{.871}	&{.895}	&{.803}	&{.045}	&.915	&{.933}	&{.880}	&\textbf{\color{blue}.027}	&{.868}	&\textbf{\color{blue}.896}	&{.799}	&{.043}
\\
\multicolumn{2}{c}{LSNet \cite{LSNet}}	&\textbf{\color{blue}.877} &\textbf{\color{red}.911}	&\textbf{\color{red}.827} 	&\textbf{\color{red}.033}	&\textbf{\color{blue}.924}	&\textbf{\color{blue}.936}	&\textbf{\color{red}.887}	&\textbf{\color{red}.022} &\textbf{\color{blue}.876}	&\textbf{\color{red}.916}	&\textbf{\color{blue}.827}	&\textbf{\color{red}.036}
\\
\multicolumn{2}{c}{DPLNet (Ours)}
&\textbf{\color{red}.878} &\textbf{\color{blue}.908}	&\textbf{\color{blue}.810} 	&\textbf{\color{blue}.043}	&\textbf{\color{red}.928}	&\textbf{\color{red}.951}	&\textbf{\color{blue}.881}	&\textbf{\color{red}.022 } &\textbf{\color{red}.879}	&\textbf{\color{red}.916}	&\textbf{\color{red}.828}	&\textbf{\color{blue}.038}
\\
\hline
\end{tabular}
\caption{\centering Results and comparison on RGB-T SOD benchmarks. ${\uparrow }/ {\downarrow}$ indicates that a larger/smaller value is better.}
\label{tb:9}
\end{table*}


\subsection{Ablation Study}
\label{4.3}
We ablate our proposed DPLNet with different experiment settings on NYUDv2 benchmark \cite{nyuv2} and all the ablation studies are conducted with single-scale inference.

\vspace{0.2em}
\noindent{\textbf{Ablation on MPG.}} To validate the effectiveness of our MPG, we replace this module with additive fusion at each stage (please notice that we keep the patch embedding layer). As in Tab.~\ref{tb:5}, the performance degrades by 0.9\% in mIoU without MPG, which demonstrates the necessity of MPG. We further ablate on the specific position of MPG module. As shown in Tab.~\ref{tb:6(a)} and Tab.~\ref{tb:6(b)}, the results indicate that using all MPG modules can achieve the best performance.

\vspace{0.2em}
\noindent{\textbf{Ablation on MFA.}} Tab.~\ref{tb:5} shows the ablation study on the proposed MFA module. We directly remove MFA at each stage and the mIoU drops by 0.9\%. Besides, we ablate the length and dimension of learnable prompt tokens $H_{k}^{i}$ in Tab.~\ref{tb:6(c)} and Tab.~\ref{tb:6(d)}, respectively. We observe that choosing 30 tokens and setting the prompt dimension to 32 lead to the best performance. Moreover, we ablate the position and numbers of MFA in DPLNet. Similar to ablation of MPG, we adopt bottom to top and top to bottom approaches as in Tab.~\ref{tb:6(e)} and in Tab.~\ref{tb:6(f)}, respectively. As shown, we observe that the result is improved when using MFA in all stages.

\vspace{0.2em}
\noindent{\textbf{Impact of different training paradigm.}} We further assess our method on two different training paradigms: (i) freezing segmentation decoder and (ii) fully fine-tuning the whole network. As shown in Tab.~\ref{tb:5}, the performance drops by 3.2\% in mIoU when we freeze the segmentation decoder. We argue that this is caused by the domain gap in the complex dense prediction tasks. When fully fine-tuning the whole network, we observe from Tab.~\ref{tb:5} that our DPLNet with 0.583 mIoU score achieves better performance than the fully fine-tuned version with 0.581 mIoU score while satisfying parameter-efficiency, which clearly shows the effectiveness and efficiency of our method.


\subsection{Generalization to Other Multimodal Tasks}
\label{4.4}
In order to show the generality of DPLNet, we conduct experiments on other multimodal segmentation tasks including RGB-T video semantic segmentation, RGB-D salient object detection, and RGB-T salient object detection.

\vspace{0.2em}
\noindent{\textbf{RGB-T video semantic segmentation.}} We assess DPLNet on RGB-T video semantic segmentation benchmark \cite{MVNet}. As shown in Tab.~\ref{tb:7}, our method achieves the best result with 0.579 mIoU, outperforming the second best MVNet with 0.545 mIoU while reducing the trainable parameters by 12$\times$. Moreover, it is worth noting that we only use the current frame to handle video semantic segmentation tasks instead of employing temporal frames as in~\cite{MVNet}, which demonstrates that DPLNet is general and applicable to multimodal video semantic segmentation.

\vspace{0.2em}
\noindent{\textbf{RGB-D salient object detection.}} For RGB-D salient object detection, we compare our DPLNet with existing competitive methods across five datasets (NJU2K \cite{(NJU2K}, NLPR \cite{NLPR}, DES \cite{DES}, SIP \cite{SIP}, and LFSD \cite{LFSD}). For performance evaluation, we follow LSNet \cite{LSNet} and adopt four metrics (\textit{e.g.}, Structure-measure (S) \cite{structuremeature}, Mean Absolute Error (M) \cite{MAEmeasure}, F-measure (F) \cite{Fmeasure} and E-measure (E) \cite{Emeasure}) for this task. As shown in Tab.~\ref{tb:8}, our method achieves SOTA performance on most of the metrics.

\noindent{\textbf{RGB-T salient object detection.}} Moreover, we evaluate DPLNet on three RGB-T salient object detection datasets \cite{VT821, VT1000, VT5000}, following settings and results in~\cite{LSNet}. As in Tab.~\ref{tb:9}, our DPLNet achieves SOTA and competitive performance on these benchmarks. The method of LSNet shows better results than DPLNet on some of the metrics. However, it adopts complicated multi-loss supervision to select features, while our DPLNet only employs a simple cross-entropy loss, without any special design in architecture, which makes it more easy to be generalized to other tasks.

\section{Conclusion}
In this paper, we introduce DPLNet, a surprisingly simple yet effective framework for training-efficient multimodal semantic segmentation. In our DPLNet, we propose a multimodal prompt generator (MPG) module to fuse different modalities, and present a multimodal feature adapter (MFA) module to adapt the frozen pre-trained backbone for better multimodal feature extraction. Our method achieves SOTA or comparable performance on RGB-D/T semantic segmentation. Moreover, DPLNet can be easily generalized and applied to other multimodal tasks such as video object detection and salient object detection. By proposing DPLNet, we hope it inspires more future work on exploring training-efficient multimodal dense prediction tasks. 

{
\small
\bibliographystyle{ieeenat_fullname}
\bibliography{main}
}

\end{document}